\documentclass[oneside,11pt]{article}

\usepackage[utf8]{inputenc}
\usepackage{amsmath}
\usepackage{amsfonts}
\usepackage{natbib}
\usepackage{subfig}
\usepackage{algorithm}
\usepackage{algpseudocode}
\inputencoding{utf8}

\usepackage{hyperref,color,soul,booktabs}
\setulcolor{blue}
\newcommand{\XXX}{\mathbb{X}}
\newcommand{\CH}{\rm CH}
\newcommand{\M}{\mathcal M}
\newcommand{\proj}[1]{^{\downarrow #1}}
\newcommand{\ext}{\rm ext}
\newcommand{\ind}{\perp \hspace{-0.6em} \perp}

\newtheorem{theorem}{Theorem}[section]
\newtheorem{lemma}[theorem]{Lemma}

\newtheorem{definition}[theorem]{Definition}
\newtheorem{example}[theorem]{Example}

\newenvironment{proof}[1][Proof]{\begin{trivlist}
\item[\hskip \labelsep {\bfseries #1}]}{\end{trivlist}}

\newcommand{\qed}{\nobreak \ifvmode \relax \else
      \ifdim\lastskip<1.5em \hskip-\lastskip
      \hskip1.5em plus0em minus0.5em \fi \nobreak
      \vrule height0.75em width0.5em depth0.25em\fi}

\usepackage{authblk}

\title{Composition of Credal Sets via Polyhedral Geometry}

\author[1]{Ji\v{r}ina Vejnarov\'{a}\thanks{vejnar@utia.cas.cz}}
\author[1]{V\'{a}clav Kratochv\'{i}l\thanks{velorex@utia.cas.cz}}
\affil[1]{Institute of Information Theory and Automation, \\
Czech Academy of Sciences, Prague (Czech Republic)}

\begin{document}

\maketitle
\begin{abstract}
Recently introduced composition operator for credal sets is an analogy of such operators in probability, possibility, evidence and valuation-based systems theories. It was designed to construct multidimensional models (in the framework of credal sets) from a system of low-dimensional credal sets. In this paper we study its potential from the computational point of view utilizing methods of polyhedral geometry.
\end{abstract}

\section{Introduction}
In the second half of 1990's a new approach to efficient
representation of multidimensional probability distributions was
introduced with the aim to be an alternative to Graphical Markov
Modeling. This approach is based on a simple idea: a
multidimensional distribution is \emph{composed} from a system of
low-dimensional distributions by repetitive application of a special
composition operator, which is also the reason why such models are
called \emph{compositional models}.

Later, these compositional models were introduced also in
possibility theory \citep{ipmu,kyber} (here the models are
parametrized by a continuous $t$-norm) and ten years ago
also in evidence theory \citep{ISIPTA07,IJAR}. In all these
frameworks the original idea is kept, but there exist some slight
differences among these frameworks.

In \citep{isipta13} we introduced a composition operator for credal
sets, but due to the problem of discontinuity it needed a revision.
After a thorough reconsideration in \citep{smps} we presented a new
definition avoiding this discontinuity. We showed that the revised
composition operator keeps the basic
properties of its counterparts in other frameworks, and therefore it
enables us to introduce compositional models for multidimensional
credal sets. Nevertheless, a problem how to make practical
computations appeared and the need for effective computational procedures became urgent.

Credal sets are usually defined as convex sets of probability distributions. In finite case, a probability distribution can be represented as a point in a multidimensional space. Credal set --- as a convex set of such points --- can be interpreted as a convex polyhedron in respective space. This naturally leads to computational procedures based on methods used in polyhedral geometry \citep{grunbaum1967convex}.

This contributions is organized as follows. In Section~2 we summarise the
basic concepts and notation. The definition of the operator of
composition is presented in Section~3, which is devoted also to its
basic properties. In Section~4 we describe proposed computational procedures, in Section 5 we illustrate their application on a few simple examples and Section~6 is devoted to implementation.

\section{Basic Concepts and Notation}

In this section we will briefly recall basic concepts and notation
necessary for understanding the contribution.

\subsection{Variables and Distributions}

For an index set $N = \{1, 2, \ldots , n\}$ let $\{X_i\}_{i \in N}$
be a system of variables, each $X_i$ having its values in a finite
set $\XXX_i$ and $ \XXX_N = \XXX_1 \times \XXX_2 \times \ldots
\times \XXX_n$ be the Cartesian product of these sets.

In this paper we will deal with groups of variables on its
subspaces. Let  $X_K$ will denote a group of variables
$\{X_i\}_{i \in K}$ with values in $\XXX_K = \mbox{\LARGE
$\times$}_{i\in K} {\XXX}_i$ throughout the paper.

Any group of variables $X_K$ can be described by a {\em probability
distribution} (sometimes also called {\em probability function})
$$P: \XXX_K \longrightarrow [0,1],$$ such that $$\sum_{x_K \in \XXX_K}
P(x_K) = 1.$$

Having two probability distributions $P_1$ and $P_2$ of $X_K$ we say
that $P_1$ is {\it absolutely continuous} with respect to $P_2$ (and
denote $P_1 \ll P_2$) if for any $x_K \in \XXX_K$
$$P_2(x_K)= 0  \Longrightarrow P_1(x_K)=0.$$
This concept plays an important role in the definition of the
composition operator.

\subsection{Credal Sets}

A {\em credal set} ${\cal M}(X_K)$ describing a group of variables
$X_K$ is usually defined as a closed convex set of probability
measures describing the values of these variables. In order to
simplify the expression of operations with credal sets, it is often
considered \citep{moral} that a credal set is the set of probability
distributions associated to the probability measures in it. Under
such consideration a credal set can be expressed as a {\it convex
hull} (denoted by CH) of its extreme distributions (ext)
$${\cal M} (X_K) = \CH \{ext({\cal M}({\it X_K}))\}.$$

Consider a credal set ${\cal M} (X_K)$. For each $L \subset K$ its {\em
marginal credal set\/} ${\cal M} (X_L)$ is obtained by element-wise
marginalization, i.e.
\begin{equation}\label{eq:margi}
{\cal M} (X_L) = {\CH} \{P\proj L: P \in {\ext}({\cal M}(X_K))\},
\end{equation}
where $P\proj L$ denotes the marginal distribution of $P$ on
$\XXX_L$. 

Besides marginalization we will also need the opposite operation, usually
called extension. {\it Vacuous extension} of a credal set
${\cal M} (X_L)$ describing $X_L$ to a credal set ${\cal M} (X_K) =
{\cal M} (X_L)^{\uparrow K}$ $(L \subset K$) is the maximal credal
set describing $X_K$ such that ${\cal M} (X_K) \proj L = {\cal M}
(X_L).$ A simple example of vacuous extension can be found in Section~\ref{sec:examples} (Example~\ref{ex:b}).

Having two credal sets ${\cal M}_1$ and ${\cal M}_2$ describing
$X_K$ and $X_L$, respectively (assuming that $K, L \subseteq N$), we
say that these credal sets are \emph{projective} if their marginals
describing common variables coincide, i.e. if $${\cal M}_1 (X_{K
\cap L}) = {\cal M}_2 (X_{K \cap L}).$$

\noindent Let us note that if $K$ and $L$ are disjoint, then ${\cal
M}_1$ and ${\cal M}_2$ are always projective, as ${\cal
M}_1(X_\emptyset)= {\cal M}_2(X_\emptyset)~\equiv~1$.

\subsection{Strong Independence}

Among the numerous definitions of independence for credal sets
\citep{couso} we have chosen strong independence, as it seems to be
the most appropriate for multidimensional models.

We say that (groups of) variables $X_K$ and $X_L$ ($K$ and $L$
disjoint) are {\it strongly independent} with respect to ${\cal
M}(X_{K\cup L})$ iff (in terms of probability distributions)
\begin{eqnarray}\label{eq:strong}
{\cal M}(X_{K\cup L})
& = & {\CH} \{P_1 \cdot P_2: P_1 \in {\cal M}(X_K), P_2 \in {\cal
M}(X_L)\}.
\end{eqnarray}

Again, there exist several generalizations of this notion to
conditional independence, see e.g. \citep{moral}, but as the
following definition is suggested by the authors as the most
appropriate for the marginal problem, it seems to be a suitable
concept also in our case, since the operator of composition can also
be used as a tool for solution of a marginal problem, as shown (in
the framework of possibility theory) e.g. in \citep{kyber}.

Given three groups of variables $X_K, X_L$ and $X_M$ ($K, L, M$ be
mutually disjoint subsets of $N$, such that $K$ and $L$ are
nonempty), we say that $X_K$ and $X_L$ are {\em conditionally independent} 
given $X_M$ under global set ${\cal M} (X_{K \cup L \cup M})$ (to
simplify the notation we will denote this relationship by $K \ind L
| M $) iff
\begin{eqnarray}
\lefteqn{ {\cal M} (X_{K \cup L \cup M})}\\ &&= {\CH} \{(P_1 \cdot
P_2)/P_1^{\proj M}: P_1 \in {\cal
M}(X_{K \cup M}),
P_2 \in {\cal M}(X_{L \cup M}), P_1 \proj M=P_2 \proj M\}\,.\nonumber
\end{eqnarray}

This definition is a generalisation of stochastic conditional
independence: if ${\cal M} (X_{K \cup L \cup M})$ is a singleton,
then ${\cal M} (X_{K \cup M})$ and ${\cal M} (X_{L \cup M})$ are
also (projective) singletons and the definition reduces to the
definition of stochastic conditional independence.

\subsection{Polyhedral Geometry}

A convex polytope may be defined in numerous ways, depending on what is more suitable for the problem at hand. Grünbaum's definition \citep{grunbaum1967convex} is in terms of a convex set of points in space. Other important definitions are: as the intersection of half-spaces (H-representation)  and as the convex hull of a set of vertices (V-representation). For a compact convex polytope, the minimal V-representation is unique and it is given by the set of the vertices of the polytope \citep{grunbaum1967convex}. In our experiments, we use both H and V-representations. However, for the purpose of representation of the polytope within the paper only V-representation is used.

As mentioned in the introduction, a credal set is a convex polytope (bounded polyhedron) in $|\XXX_N|$-dimensional space. Each dimension corresponds to an element from $\XXX_N$ (a combination of the variables $N$) - i.e. $x_N \in \XXX_N$. Vertex $v$ of the space is nothing else than a probability distribution $P_v$ with $v[x_N] = P_v(x_N)$. Following the fact, that probability belongs to unit interval $[0,1]$, we can restrict the space --- it is, in fact, a hypercube.

The nature of the space imposes several restrictions to operations with convex polytopes.
We do not consider projections into subspaces. Marginalization of a convex polytope representing a set of probability distributions of $X_N$ to $X_K$ ($K\subset N$) corresponds to a transformation of $|\XXX_N|$-dimensional space into $\XXX_K$-dimensional space. Naturally, for each vertex this is done by summing of all coordinates from $\XXX_N$ with the projection to $\XXX_K$. The similar holds for extensions.

\section{Composition Operator}

In this section we will recall the new definition of composition
operator 
 for credal sets introduced in \citep{smps}. 
To enable the reader better
understanding to the concept we will  present it first in a
precise probability framework.

\subsection{Composition Operator of Probability Distributions}

First let us recall the definition of composition of two probability
distributions \citep{uai97}. Consider two index sets $K, L \subset
N$. We do not put any restrictions on $K$ and $L$; they may be but
need not be disjoint, and one may be a subset of the other. Let
$P_1$ and $P_2$ be two probability distributions of (groups of)
variables $X_K$ and $X_L$; then
\begin{equation}\label{eq:opprob}
(P_1 \triangleright P_2) (X_{K \cup L}) = \frac{P_1(X_K) \cdot
P_2(X_L)}{P_2(X_{K \cap L})},
\end{equation}
whenever $P_1(X_{K \cap L}) \ll P_2(X_{K \cap L})$; otherwise, it
remains undefined.

It is specific property of composition operator for probability
distributions --- in other settings the operator is always defined
\citep{kyber,ISIPTA07}.

\subsection{Definition}

Let ${\cal M}_1$ and ${\cal M}_2$ be credal sets describing $X_K$
and $X_L$, respectively. Our goal is to define a new credal set,
denoted by ${\cal M}_1 \triangleright {\cal M}_2$, which will describe
$X_{K \cup L}$ and will contain all of the information
contained in ${\cal M}_1$ and, as much as possible, from ${\cal M}_2$. In other words, we want to find a common extension of ${\cal M}_1$ and ${\cal M}_2$ (if it is possible).

The required properties were already met by Definition~1 in
\citep{isipta13}\footnote{Let us note that the definition is based on
Moral's concept of conditional independence with relaxing
convexity.}. However, that definition exhibits a kind of
discontinuity and was thoroughly reconsidered. In \citep{smps} we
 proposed the following one.

\begin{definition} \label{def:operator}
{\rm For two credal sets ${\cal M}_1$ and ${\cal M}_2$ describing
$X_K$ and $X_L$, their \emph{composition} ${\cal M}_1 \triangleright
{\cal M}_2$ is defined as a convex hull of probability distributions
$P$ obtained as follows. For each couple of distributions $P_1 \in
{\cal M}_1(X_{K})$ and $P_2 \in{\cal M}_2(X_{L})$ such that $P_2
\proj{K \cap L} \in argmin \{Q_2 \in {\cal M}_2(X_{K \cap L}): d(Q_2
,P_1 \proj{K \cap L})$, distribution $P$ is obtained by one of the
following rules:
\begin{description}
\item[[a\hspace{-1ex}]] if $P_1\proj{K \cap L}
\ll P_2\proj{K \cap L}$
$$P(X_{K \cup L}) = \frac{P_1 (X_{K})\cdot P_2 (X_{L})}{P_2 \proj{K \cap L}(X_{K \cap L})}.$$
\item[[b\hspace{-1ex}]] otherwise
$$P (X_{K \cup L}) \in {\ext}\{P_1^{\uparrow K \cup L}(X_{K})\}.$$
\end{description}
 }\end{definition}

Function $d$ used in the definition is a suitable distance function.
In this paper we use Euclidean distance, as it is natural choice in polyhedral geometry.

Let us note, that this definition of composition operator does not
differ from the original one \citep{isipta13} in case of projective
credal sets, as in this case the only distributions in ${\cal M}_1
\triangleright {\cal M}_2$ are those satisfying $P = (P_1 \cdot
P_2)/P_2 \proj{K \cap L}$, where $P_1 \proj{K \cap L} = P_2 \proj{K
\cap L}$ (and those belonging to their convex hull). However, it differs in the remaining cases. It will be
illustrated in Section~5 using our computational procedures.

In the next subsection we will summarize the most important basic properties of the composition operator.

\subsection{Basic Properties}

The following lemma proven in \citep{smps} suggests that the
above-defined composition operator possesses basic properties
required at the beginning of this section. Its last item characterizes condition under which common extension of ${\cal M}_1$ and ${\cal M}_2$ can be obtained.

\begin{lemma}
\label{basiclemma1} For two credal sets ${\cal M}_1$ and ${\cal
M}_2$ describing $X_K$ and $X_L$, respectively, the following
properties hold true:
\begin{enumerate}
\item ${\cal M}_1 \triangleright {\cal M}_2$ is a credal set describing $X_{K \cup
L}$.\label{bp0}
\item $({\cal M}_1 \triangleright {\cal M}_2) (X_K) = {\cal M}_1 (X_K)$.
\label{bp1}
\item ${\cal M}_1 \triangleright {\cal M}_2 = {\cal M}_2 \triangleright {\cal M}_1$ iff ${\cal M}_1 (X_{K
\cap L}) = {\cal M}_2 (X_{K \cap L}).$\label{bp2}
\end{enumerate}
\end{lemma}


This lemma, together with the following theorem, proven in \citep{isipta13}, expressing  the
relationship between strong independence and the operator of
composition, are the most important
assertions enabling us to introduce compositional models.

\begin{theorem}\label{veta:margi}
Let ${\cal M}$ be a credal set describing $X_{K \cup L}$ with
marginals ${\cal M} (X_K)$ and ${\cal M} (X_L)$. Then
$${\cal M}(X_{K \cup L}) = ({\cal M}\proj K \triangleright {\cal
M}\proj L) (X_{K \cup L})$$ iff
$$(K \setminus L )\ind (L \setminus K)| (K \cap L).$$
\end{theorem}

This theorem remains valid also for the revised definition of the
composition operator, as ${\cal M} (X_K)$ and ${\cal M} (X_L)$ are
marginals of ${\cal M}(X_{K \cup L})$, and therefore only [a] (for
projective distributions) is applicable.

Before closing this section, let us present one more result proven in \citep{smpsIJAR} concerning the relationship between the original composition operator for precise probabilities and that studied in this contribution.

\begin{lemma}
\label{lemma_probabilistic} Let ${\cal M}_1 (X_{K})$ and ${\cal M}_2
(X_{L})$ be two singleton credal sets describing $X_K$ and $X_L$,
respectively, where ${\cal M}_1(X_{K \cap L})$ is absolutely
continuous with respect to ${\cal M}_2(X_{K \cap L})$.  Then $({\cal
M}_1 \triangleright {\cal M}_2)(X_{K \cup L})$ is also a singleton.
\end{lemma}

The reader should however realize that the definition of the
operator of composition for singleton credal sets is not completely
equivalent to the definition of composition for probabilistic
distributions. They equal each other only in case that the
probabilistic version is defined. This is ensured in
Lemma~\ref{lemma_probabilistic} by assuming the absolute continuity.
In case it does not hold, the probabilistic operator is not defined
while its credal version introduced in this paper is always defined. However, in
this case, the result is not a singleton credal set, as can be seen from Example~\ref{ex:b} in Section~\ref{sec:examples}.

\section{Computational Procedures for Composition Operator}
The experiments were performed in R environment \citep{Rproject} using RStudio \citep{Rstudio}. To implement polyhedral geometry, we have used \emph{rcdd} package \citep{rcdd} based on GMP GNU library (The GNU Multiple Precision Arithmetic Library) \citep{Granlund12}.

For quadratic programming methods --- i.e. in case of a finding a projection of a given vertex on a certain polytope (we considered Euclidean distance in this paper) ---  we have used methods of quadratic programming \citep{gould2000quadratic} implemented in \emph{quadprog} package \citep{quadprog}.

To implement various extensions to higher dimensional space --- as described in
Definition~\ref{def:operator} --- we have used the advantage of equivalence of H- and V-representations of
convex polytopes. To move from $|\XXX_N|$-dimensional space to $|\XXX_K|$-dimensional space where
$K \subset N$ we can easily create a 0-1 transformation matrix that allows us to convert an arbitrary
vertex from one space to the other one. With the help of such a transformation matrix, a set of related
(in)equalities can be transformed as well. This is extremely handy when we need to extend a vertex or a
convex polytope from a space to the higher-dimensional one. First, we convert it to its H-representation
and then transform it using a transformation matrix to the higher dimensional space.

To find a part of a polytope with given projection, it is enough to extend H-projection of the projection to the original space and combine both H-representations (to get the intersection of the polytopes).

\section{Examples}\label{sec:examples}
In this Section we will demonstrate the application of Definition~1 via our computational procedures. Let us start with the case of projective credal sets.

\begin{example}\label{ex:1}

Let ${\cal M}_1 (X_1 X_2)$ and ${\cal M}_2 (X_2
X_3)$ be credal sets about variables $X_1X_2$ and $X_2X_3$, respectively, with extreme vertices as listed in Table \ref{tbl:ex1}.

\begin{table}[!htb]
    \begin{minipage}{.5\linewidth}
      \centering
\begin{tabular}{lllll}
  \hline
 & $x_1x_2$ & $x_1\bar{x_2}$ & $\bar{x_1}x_2$ & $\bar{x_1}\bar{x_2}$ \\
  \hline
1 & 0.2 & 0.2 & 0 & 0.6 \\
  2 & 0.1 & 0.4 & 0.1 & 0.4 \\
  3 & 0.25 & 0.25 & 0.25 & 0.25 \\
  4 & 0.2 & 0.3 & 0.3 & 0.2 \\
   \hline
\end{tabular}
      \caption*{${\cal M}_1 (X_1 X_2)$}
    \end{minipage}%
    \begin{minipage}{.5\linewidth}
      \centering
\begin{tabular}{lllll}
  \hline
 & $x_2x_3$ & $x_2\bar{x_3}$ & $\bar{x_2}x_3$ & $\bar{x_2}\bar{x_3}$ \\
  \hline
1 & 0.2 & 0 & 0.3 & 0.5 \\
  2 & 0 & 0.2 & 0 & 0.8 \\
  3 & 0.5 & 0 & 0.5 & 0 \\
  4 & 0.2 & 0.3 & 0.2 & 0.3 \\
   \hline
\end{tabular}
\caption*{${\cal M}_2 (X_2 X_3)$}
    \end{minipage}

\caption{V-representations of credal sets ${\cal M}_1$ and ${\cal M}_2$ from Example \ref{ex:1}}
\label{tbl:ex1}
\end{table}

These two credal sets are projective, as
$${\cal M}_1 (X_2)= \CH \{[0.2,0.8], [0.5, 0.5]\}={\cal M}_2
(X_2)$$ Following Definition~1, ${\cal M}_1
\triangleright {\cal M}_2$  can be expressed as a convex polytope with V-representation defined in Table \ref{tbl:ex1out}.

\begin{table}[ht]
\centering
\begin{tabular}{lllllllll}
  \hline
 & $x_1x_2x_3$ & $x_1x_2\bar{x_3}$ & $x_1\bar{x_2}x_3$ & $x_1\bar{x_2}\bar{x_3}$ & $\bar{x_1}x_2x_3$ & $\bar{x_1}x_2\bar{x_3}$ & $\bar{x_1}\bar{x_2}x_3$ & $\bar{x_1}\bar{x_2}\bar{x_3}$ \\
  \hline
1 & 0 & 0.2 & 0 & 0.8 & 0 & 0 & 0 & 0 \\
  2 & 0.13 & 0.07 & 0.46 & 0.34 & 0 & 0 & 0 & 0 \\
  3 & 0 & 0.1 & 0 & 0.4 & 0 & 0.1 & 0 & 0.4 \\
  4 & 0.07 & 0.03 & 0.23 & 0.17 & 0.07 & 0.03 & 0.23 & 0.17 \\
  5 & 0.3 & 0 & 0.2 & 0 & 0.3 & 0 & 0.2 & 0 \\
  6 & 0.15 & 0.15 & 0.1 & 0.1 & 0.15 & 0.15 & 0.1 & 0.1 \\
  7 & 0 & 0 & 0 & 0 & 0.6 & 0 & 0.4 & 0 \\
  8 & 0 & 0 & 0 & 0 & 0.3 & 0.3 & 0.2 & 0.2 \\
   \hline
\end{tabular}
\caption{V-representation of ${\cal M}_1 \triangleright {\cal M}_2$ from Example 1}
\label{tbl:ex1out}
\end{table}
It can easily be checked that both $({\cal M}_1 \triangleright {\cal
M}_2) (X_1 X_2) = {\cal M}_1 (X_1 X_2)$ and $({\cal M}_1
\triangleright {\cal M}_2) (X_2 X_3)= {\cal M}_2 (X_2
X_3)$.\hfill$\diamondsuit$
\end{example}

Here for any distribution $P_1(X_1 X_2)$ in ${\cal M}_1$ exists (at least one) distribution $P_2(X_2 X_3)$ in ${\cal M}_2$ such that $P_1(X_2) = P_2(X_2)$, therefore all the extreme points are obtained as a simple conditional product. In this case, furthermore, ${\cal M}_1
\triangleright {\cal M}_2 = {\cal M}_2 \triangleright {\cal M}_1$, as corresponds to Lemma~\ref{basiclemma1} (and can be easily checked).

The following example is more complicated, as it deals with non-projective credal sets.

\begin{example}\label{ex:2}

Let ${\cal M}_1(X_1 X_2)$ and ${\cal M}_2 (X_2 X_3)$
be two credal sets describing binary variables $X_1X_2$ and
$X_2X_3$, respectively, defined as a convex hull of vertices in Table \ref{tbl:ex2In}.

\begin{table}[!htb]
    \begin{minipage}{.5\linewidth}
      \centering
\begin{tabular}{lllll}
  \hline
 & $x_1x_2$ & $x_1\bar{x_2}$ & $\bar{x_1}x_2$ & $\bar{x_1}\bar{x_2}$ \\
  \hline
1 & 0.2 & 0.8 & 0 & 0 \\
  2 & 0.1 & 0.4 & 0.1 & 0.4 \\
  3 & 0.3 & 0.2 & 0.3 & 0.2 \\
  4 & 0 & 0 & 0.6 & 0.4 \\
   \hline
\end{tabular}
      \caption*{${\cal M}_1 (X_1 X_2)$}
    \end{minipage}%
    \begin{minipage}{.5\linewidth}
      \centering
\begin{tabular}{lllll}
  \hline
 & $x_2x_3$ & $x_2\bar{x_3}$ & $\bar{x_2}x_3$ & $\bar{x_2}\bar{x_3}$ \\
  \hline
1 & 0 & 0.3 & 0 & 0.7 \\
  2 & 0.2 & 0.1 & 0.4 & 0.3 \\
  3 & 0.25 & 0.25 & 0.25 & 0.25 \\
  4 & 0.5 & 0 & 0.5 & 0 \\
   \hline
\end{tabular}
\caption*{${\cal M}_2 (X_2 X_3)$}
    \end{minipage}

\caption{V-representations of credal sets ${\cal M}_1$ and ${\cal M}_2$ from Example \ref{ex:2}}
\label{tbl:ex2In}
\end{table}

These two credal sets are not projective, as
${\cal M}_1 (X_2)=  \CH \{[0.2,0.8], [0.6, 0.4]\},$ while ${\cal
M}_2 (X_2)=  \CH \{[0.3,0.7], [0.5, 0.5]\}.$ Therefore ${\cal M}_2
(X_2) \subset {\cal M}_1 (X_2)$.
Definition~1 in this case leads to a credal set
$({\cal M}_1 \triangleright {\cal M}_2) (X_1
X_2 X_3)$ with 23 extreme points listed in Table~\ref{tbl:ex2_12}.
\begin{table}[!htb]
\centering
\begin{tabular}{lllllllll}
  \hline
 & $x_1x_2x_3$ & $x_1x_2\bar{x_3}$ & $x_1\bar{x_2}x_3$ & $x_1\bar{x_2}\bar{x_3}$ & $\bar{x_1}x_2x_3$ & $\bar{x_1}x_2\bar{x_3}$ & $\bar{x_1}\bar{x_2}x_3$ & $\bar{x_1}\bar{x_2}\bar{x_3}$ \\
  \hline
1 & 0 & 0.15 & 0 & 0.35 & 0 & 0.15 & 0 & 0.35 \\
  2 & 0 & 0.075 & 0 & 0.3 & 0 & 0.225 & 0 & 0.4 \\
  3 & 0.05 & 0.025 & 0.171 & 0.129 & 0.15 & 0.075 & 0.229 & 0.171 \\
  4 & 0 & 0.15 & 0 & 0.6 & 0 & 0.15 & 0 & 0.1 \\
  5 & 0.1 & 0.05 & 0.343 & 0.257 & 0.1 & 0.05 & 0.057 & 0.043 \\
  6 & 0 & 0.225 & 0 & 0.65 & 0 & 0.075 & 0 & 0.05 \\
  7 & 0.15 & 0.075 & 0.371 & 0.279 & 0.05 & 0.025 & 0.029 & 0.021 \\
  8 & 0.125 & 0.125 & 0.125 & 0.125 & 0.125 & 0.125 & 0.125 & 0.125 \\
  9 & 0.25 & 0 & 0.25 & 0 & 0.25 & 0 & 0.25 & 0 \\
  1 & 0.012 & 0.012 & 0.05 & 0.05 & 0.238 & 0.238 & 0.2 & 0.2 \\
  11 & 0.025 & 0 & 0.1 & 0 & 0.475 & 0 & 0.4 & 0 \\
  12 & 0.025 & 0.025 & 0.1 & 0.1 & 0.225 & 0.225 & 0.15 & 0.15 \\
  13 & 0.05 & 0 & 0.2 & 0 & 0.45 & 0 & 0.3 & 0 \\
  14 & 0.138 & 0.138 & 0.175 & 0.175 & 0.113 & 0.113 & 0.075 & 0.075 \\
  15 & 0.275 & 0 & 0.35 & 0 & 0.225 & 0 & 0.15 & 0 \\
  16 & 0 & 0.2 & 0 & 0.8 & 0 & 0 & 0 & 0 \\
  17 & 0.133 & 0.067 & 0.457 & 0.343 & 0 & 0 & 0 & 0 \\
  18 & 0 & 0.1 & 0 & 0.4 & 0 & 0.1 & 0 & 0.4 \\
  19 & 0.067 & 0.033 & 0.229 & 0.171 & 0.067 & 0.033 & 0.229 & 0.171 \\
  2 & 0.3 & 0 & 0.2 & 0 & 0.3 & 0 & 0.2 & 0 \\
  21 & 0.15 & 0.15 & 0.1 & 0.1 & 0.15 & 0.15 & 0.1 & 0.1 \\
  22 & 0 & 0 & 0 & 0 & 0.6 & 0 & 0.4 & 0 \\
  23 & 0 & 0 & 0 & 0 & 0.3 & 0.3 & 0.2 & 0.2 \\
   \hline
\end{tabular}
\caption{V-representation $({\cal M}_1 \triangleright {\cal M}_2) (X_1
X_2 X_3)$ from Example \ref{ex:2}}.
\label{tbl:ex2_12}
\end{table}
On the other hand $({\cal M}_2 \triangleright {\cal M}_1) (X_1
X_2 X_3)$ has 16 extreme points. They are listed in Table~\ref{tbl:ex2_21}.
\begin{table}[ht]
\centering
\begin{tabular}{lllllllll}
  \hline
 & $x_1x_2x_3$ & $x_1x_2\bar{x_3}$ & $x_1\bar{x_2}x_3$ & $x_1\bar{x_2}\bar{x_3}$ & $\bar{x_1}x_2x_3$ & $\bar{x_1}x_2\bar{x_3}$ & $\bar{x_1}\bar{x_2}x_3$ & $\bar{x_1}\bar{x_2}\bar{x_3}$ \\
  \hline
1 & 0 & 0.15 & 0 & 0.35 & 0 & 0.15 & 0 & 0.35 \\
  2 & 0 & 0.075 & 0 & 0.3 & 0 & 0.225 & 0 & 0.4 \\
  3 & 0 & 0.15 & 0 & 0.6 & 0 & 0.15 & 0 & 0.1 \\
  4 & 0 & 0.225 & 0 & 0.65 & 0 & 0.075 & 0 & 0.05 \\
  5 & 0.1 & 0.05 & 0.2 & 0.15 & 0.1 & 0.05 & 0.2 & 0.15 \\
  6 & 0.05 & 0.025 & 0.171 & 0.129 & 0.15 & 0.075 & 0.229 & 0.171 \\
  7 & 0.1 & 0.05 & 0.343 & 0.257 & 0.1 & 0.05 & 0.057 & 0.043 \\
  8 & 0.15 & 0.075 & 0.371 & 0.279 & 0.05 & 0.025 & 0.029 & 0.021 \\
  9 & 0.125 & 0.125 & 0.125 & 0.125 & 0.125 & 0.125 & 0.125 & 0.125 \\
  1 & 0.012 & 0.012 & 0.05 & 0.05 & 0.238 & 0.238 & 0.2 & 0.2 \\
  11 & 0.025 & 0.025 & 0.1 & 0.1 & 0.225 & 0.225 & 0.15 & 0.15 \\
  12 & 0.138 & 0.138 & 0.175 & 0.175 & 0.113 & 0.113 & 0.075 & 0.075 \\
  13 & 0.25 & 0 & 0.25 & 0 & 0.25 & 0 & 0.25 & 0 \\
  14 & 0.025 & 0 & 0.1 & 0 & 0.475 & 0 & 0.4 & 0 \\
  15 & 0.05 & 0 & 0.2 & 0 & 0.45 & 0 & 0.3 & 0 \\
  16 & 0.275 & 0 & 0.35 & 0 & 0.225 & 0 & 0.15 & 0 \\
   \hline
\end{tabular}
\caption{V-representation of $({\cal M}_2 \triangleright {\cal M}_1) (X_1
X_2 X_3)$ from Example \ref{ex:2}}
\label{tbl:ex2_21}
\end{table}

\hfill$\diamondsuit$
\end{example}

 This difference deserves a more detailed explanation. Both ${\cal M}_1
\triangleright {\cal M}_2$ and ${\cal M}_2 \triangleright
{\cal M}_1$ keep the first marginal, but not the second one (which corresponds to Lemma~\ref{basiclemma1}).

The extreme vertices of marginal of the composition ${\cal M}_2 \triangleright
{\cal M}_1$ of $X_1X_2$ are listed in Table~\ref{tbl:ex2m1}. Is is not obvious just from the table itself, but the convex polytope corresponding to $({\cal M}_2 \triangleright {\cal M}_1) (X_1 X_2)$ is smaller than the one of the original credal set ${\cal M}_1(X_1X_2)$. Actually, any extreme vertex of $({\cal M}_2 \triangleright {\cal M}_1) (X_1X_2)$ is an inner point of ${\cal M}_1 (X_1X_2)$.

The opposite inclusion holds for the marginal of the composition ${\cal M}_1 \triangleright
{\cal M}_2$ of $X_2X_3$. The V-representation of respective credal set is given in Table \ref{tbl:ex2m2}. This credal set is bigger than ${\cal M}_2(X_2X_3)$, i.e., any extreme vertex of  ${\cal M}_2(X_2X_3)$ is contained in  $({\cal M}_1 \triangleright {\cal M}_2) (X_2X_3)$.

\begin{table}[!htb]
    \begin{minipage}{.5\linewidth}
      \centering
\begin{tabular}{lllll}
  \hline
 & $x_1x_2$ & $x_1\bar{x_2}$ & $\bar{x_1}x_2$ & $\bar{x_1}\bar{x_2}$ \\
  \hline
1 & 0.15 & 0.35 & 0.15 & 0.35 \\
  2 & 0.075 & 0.3 & 0.225 & 0.4 \\
  3 & 0.15 & 0.6 & 0.15 & 0.1 \\
  4 & 0.225 & 0.65 & 0.075 & 0.05 \\
  5 & 0.25 & 0.25 & 0.25 & 0.25 \\
  6 & 0.025 & 0.1 & 0.475 & 0.4 \\
  7 & 0.05 & 0.2 & 0.45 & 0.3 \\
  8 & 0.275 & 0.35 & 0.225 & 0.15 \\
   \hline
\end{tabular}
      \caption{$({\cal M}_2 \triangleright {\cal M}_1) (X_1 X_2)$}
      \label{tbl:ex2m1}
    \end{minipage}%
    \begin{minipage}{.5\linewidth}
      \centering
\begin{tabular}{lllll}
  \hline
 & $x_2x_3$ & $x_2\bar{x_3}$ & $\bar{x_2}x_3$ & $\bar{x_2}\bar{x_3}$ \\
  \hline
1 & 0 & 0.3 & 0 & 0.7 \\
  2 & 0.25 & 0.25 & 0.25 & 0.25 \\
  3 & 0.5 & 0 & 0.5 & 0 \\
  4 & 0 & 0.2 & 0 & 0.8 \\
  5 & 0.133 & 0.067 & 0.457 & 0.343 \\
  6 & 0.6 & 0 & 0.4 & 0 \\
  7 & 0.3 & 0.3 & 0.2 & 0.2 \\
   \hline
\end{tabular}
      \caption{$({\cal M}_1 \triangleright {\cal M}_2) (X_2 X_3)$}
         \label{tbl:ex2m2}
    \end{minipage}

\caption*{V-representation of two marginals of composition of ${\cal M}_1$ and ${\cal M}_2$ from Example \ref{ex:2}}
\end{table}

The fact that $({\cal M}_2 \triangleright {\cal M}_1)(X_1X_2)$ is smaller (more precise) than $({\cal M}_2 \triangleright {\cal M}_1))(X_1X_2)$ corresponds to the idea that we want  ${\cal M}_2 \triangleright {\cal M}_1$ to keep all the information contained in  ${\cal M}_2$. Therefore, we do not
consider those distributions from  ${\cal M}_1$ not corresponding to
any from  ${\cal M}_2$ --- although these distributions are taken into
account when composing ${\cal M}_1 \triangleright {\cal M}_2$.

The last example demonstrates the case, when [b] of Definition~1 is applied. Simultaneously, it is the case, when probabilistic composition operator remains undefined, but composition of two precise probabilities taken as singleton credal sets is defined (and the result is, naturally, a credal set).

\begin{example}\label{ex:b}

\begin{table}[!htb]
    \begin{minipage}{.5\linewidth}
      \centering
\begin{tabular}{lllll}
  \hline
 & $x_1x_2$ & $x_1\bar{x_2}$ & $\bar{x_1}x_2$ & $\bar{x_1}\bar{x_2}$ \\
  \hline
1 & 0.25 & 0.25 & 0.25 & 0.25\\
   \hline
\end{tabular}
      \caption*{${\cal M}_1 (X_1 X_2)$}
    \end{minipage}%
    \begin{minipage}{.5\linewidth}
      \centering
\begin{tabular}{lllll}
  \hline
 & $x_2x_3$ & $x_2\bar{x_3}$ & $\bar{x_2}x_3$ & $\bar{x_2}\bar{x_3}$ \\
  \hline
  1 & 0.5 & 0.5 & 0 & 0 \\
   \hline
\end{tabular}
\caption*{${\cal M}_2 (X_2 X_3)$}
    \end{minipage}

\caption{Two singleton credal sets ${\cal M}_1$ and ${\cal M}_2$ from Example \ref{ex:b}}
\label{tbl:exb}
\end{table}

Let ${\cal M}_1 (X_1 X_2)$ and ${\cal M}_2 (X_2 X_3)$ be two singleton
credal sets describing variables $X_1X_2$ and $X_2X_3$,
respectively. They are defined in Table \ref{tbl:exb}. Let us compute ${\cal M}_1 \triangleright {\cal M}_2$. As ${\cal M}_1 (X_2)= \{[0.5,0.5]\},$ while $({\cal M}_2 (X_2) =
\{[1,0]\},$ it is evident, that ${\cal M}_1$ is not absolutely
continuous with respect to  ${\cal M}_2$. Therefore, using part [b]
of Definition~1, we get the full extension of ${\cal M}_1 (X_2)= \{[0.5,0.5]\},$ to $\XXX_{1,2,3}$-dimensional space of probability distributions (i.e. to a hypercube). The V-representation of such a polytope is listed in Table \ref{tbl:exbOut}.

\begin{table}[ht]
\centering
\begin{tabular}{lllllllll}
  \hline
 & $x_1x_2x_3$ & $x_1x_2\bar{x_3}$ & $x_1\bar{x_2}x_3$ & $x_1\bar{x_2}\bar{x_3}$ & $\bar{x_1}x_2x_3$ & $\bar{x_1}x_2\bar{x_3}$ & $\bar{x_1}\bar{x_2}x_3$ & $\bar{x_1}\bar{x_2}\bar{x_3}$ \\
  \hline
1 & 0.25 & 0 & 0 & 0.25 & 0 & 0.25 & 0 & 0.25 \\
  2 & 0.25 & 0 & 0 & 0.25 & 0 & 0.25 & 0.25 & 0 \\
  3 & 0.25 & 0 & 0 & 0.25 & 0.25 & 0 & 0 & 0.25 \\
  4 & 0.25 & 0 & 0 & 0.25 & 0.25 & 0 & 0.25 & 0 \\
  5 & 0.25 & 0 & 0.25 & 0 & 0.25 & 0 & 0 & 0.25 \\
  6 & 0.25 & 0 & 0.25 & 0 & 0.25 & 0 & 0.25 & 0 \\
  7 & 0.25 & 0 & 0.25 & 0 & 0 & 0.25 & 0.25 & 0 \\
  8 & 0.25 & 0 & 0.25 & 0 & 0 & 0.25 & 0 & 0.25 \\
  9 & 0 & 0.25 & 0.25 & 0 & 0.25 & 0 & 0.25 & 0 \\
  10 & 0 & 0.25 & 0.25 & 0 & 0.25 & 0 & 0 & 0.25 \\
  11 & 0 & 0.25 & 0.25 & 0 & 0 & 0.25 & 0.25 & 0 \\
  12 & 0 & 0.25 & 0.25 & 0 & 0 & 0.25 & 0 & 0.25 \\
  13 & 0 & 0.25 & 0 & 0.25 & 0.25 & 0 & 0.25 & 0 \\
  14 & 0 & 0.25 & 0 & 0.25 & 0.25 & 0 & 0 & 0.25 \\
  15 & 0 & 0.25 & 0 & 0.25 & 0 & 0.25 & 0.25 & 0 \\
  16 & 0 & 0.25 & 0 & 0.25 & 0 & 0.25 & 0 & 0.25 \\
   \hline
\end{tabular}
\caption{V-representation of ${\cal M}_1 (X_2)^{\uparrow X_1X_2X_3}$}
\label{tbl:exbOut}
\end{table}
 \hfill$\diamondsuit$
\end{example}
\section{Implementation}

\begin{algorithm}
\begin{algorithmic}

\Procedure{V}{$\cal M$}
    \State \Return V-representation of $\cal M$ - set of extreme vertices
\EndProcedure

\Procedure{compose}{${\cal M}_1(X_K), {\cal M}_2(X_L)$}
    \State ${\cal M}_{KL} \gets {\cal M}_1^{K\cap L} \cap {\cal M}_2^{K\cap L}$
    \State ${\cal M}_1^{projective} \gets \M_{KL}^{\uparrow K} \cap \M_1 $
    \State ${\cal M}_2^{projective} \gets \M_{KL}^{\uparrow L} \cap \M_2 $
    \State result $ \gets \emptyset$
    \For{$P_1 \in V(\M_1^{projective})$}
        \For{$P_2 \in V(\M_2^{projective})$}
            \If{$P_1^{\downarrow K \cap L} = P_2^{\downarrow K \cap L}$}
                \State add $P_1 \triangleright P_2$ to the result
            \EndIf
        \EndFor
    \EndFor
    \For{$P_1 \in V(\M_1)$}
        \State $Q_2 \gets$ find a projection of $P_1^{\downarrow K \cap L}$ on $\M_2^{\downarrow K \cap L}$
        \State (comment: i.e. minimize distance $\lVert P_1^{\downarrow K \cap L} - p \rVert$ subject to $p \in \M_2^{\downarrow K \cap L}$)
        \State (comment: in case of Euclidean distance we can use methods of quadratic programming)
        \If{$P_1^{\downarrow K \cap L} \ll Q_2$}
            \For{$P_2 \in V(\M_2 \cap Q_2^{\uparrow L})$}
                  \State add $P_1 \triangleright P_2$ to the result
            \EndFor
        \EndIf
        \If{$P_1^{\downarrow K \cap L} \not\ll Q_2$}
            \State add $V(P_1^{\uparrow K\cup L})$  to the result
        \EndIf
    \EndFor
    \State \Return convex hall of the result
\EndProcedure
\end{algorithmic}
\caption{Implementation of ${\cal M}_1 \triangleright {\cal M}_2$}
\label{alg:1}
\end{algorithm}

The implementation of ${\cal M}_1 \triangleright {\cal M}_2$ via Definition~1 is based on a conjecture that it is sufficient to deal with two finite groups of vertices only. The first group coincides with the set of extreme vertices of each polytope (credal set). The other group corresponds to projective parts of the credal sets - i.e. the parts whose marginals describing common variables coincide. For better explanation see the pseudo-code of the implementation, as described in Algorithm \ref{alg:1}.

\section{Conclusions and Future Work}

We have presented computational procedures for composition of credal sets. We utilized the fact, that a credal set can be viewed as a special case of a convex polyhedron and that the methods in polyhedral geometry are developed for a long period. It seems to be useful, as the composition can hardly be performed without this computational support (with the exception of the simplest examples).

Nevertheless, it is only the first step in the construction of multidimensional models. The repetitive application of the composition operator is theoretically solved for so-called perfect sequences of credal sets, but the computational issues have not been tackled yet. And the problems connected with other generating sequences could be Another research direction can be shift of the distance from the Euclidean on to some divergence of probability distributions as e.g. Kullback-Leibler divergence, total variation or some other
f-divergence \citep{vajda}.

Last but not least, a number of theoretical issues concerning relationship between credal compositional models and other kinds of multidimensional models in the framework of credal sets or compositional models in other frameworks are to be solved.

\section*{Acknowledgement}
This work was supported by the Czech Science Foundation (project 16-12010S).

\vskip 0.2in
\bibliographystyle{plainnat}

\end{document}